\documentclass[10pt,twocolumn,letterpaper]{article}

\usepackage{cvpr}
\usepackage{times}
\usepackage{epsfig}
\usepackage{graphicx}
\usepackage{amsmath}
\usepackage{amssymb}
\usepackage{multirow}
\usepackage{url}
\usepackage[breaklinks=true,bookmarks=false]{hyperref}

\cvprfinalcopy % *** Uncomment this line for the final submission

\def\cvprPaperID{1294} % *** Enter the CVPR Paper ID here

\ifcvprfinal\fi

\begin{document}

%%%%%%%%% TITLE
\title{Automatic Dataset Augmentation}

\author{Yalong Bai\thanks{This work was done when the first author was an intern at Microsoft Research Asia.}\\
	%School of Computer Science and Technology\\
	Harbin Institute of Technology\\
	Harbin, 150001, P. R. China\\
	{\tt\small ylbai@mtlab.hit.edu.cn}
	\and
	Kuiyuan Yang\\
	DeepMotion\\
	No. 9 North 4th Ring West Road\\
	Beijing, 100190, P. R. China\\
	{\tt\small kuiyuanyang@deepmotion.ai}
	\and
    Tao Mei\\
    JD Research\\
    No. 8 Beichen West Street\\
    Beijing 100105, P. R. China\\
    {\tt\small tmei@jd.com}
    \and
	Wei-Ying Ma\\
	Bytedance\\
	No. 43 Beisanhuan West Road\\
	Beijing, 100080, P.R. China\\
	{\tt\small weiyingma@bytedance.com}
	\and
	Tiejun Zhao\\
	%School of Computer Science and Technology\\
	Harbin Institute of Technology\\
	Harbin, 150001, P. R. China\\
	{\tt\small tjzhao@hit.edu.cn}
}

\maketitle
\begin{abstract}
Large scale image dataset and deep convolutional neural network (DCNN) are two primary driving forces for the rapid progress made in generic object recognition tasks in recent years. While lots of network architectures have been continuously designed to pursue lower error rates, few efforts are devoted to enlarge existing datasets due to high labeling cost and unfair comparison issues. In this paper, we aim to achieve lower error rate by augmenting existing datasets in an automatic manner. Our method leverages both Web and DCNN, where Web provides massive images with rich contextual information, and DCNN replaces human to automatically label images under guidance of Web contextual information. Experiments show our method can automatically scale up existing datasets significantly from billions web pages with high accuracy, and significantly improve the performance on object recognition tasks by using the automatically augmented datasets, which demonstrates that more supervisory information has been automatically gathered from the Web. Both the dataset and models trained on the dataset are made publicly available.
\end{abstract}

\section{Introduction}
Generic object recognition is a fundamental problem in computer vision, and has achieved steady progress with efforts from both large scale dataset construction and sophisticated model design. Though the goal is to minimize expected error on previously unseen images, only empirical error can be directly optimized on a set of labeled images with respect to a function space defined by a model. According to statistical learning theory, the gap between expected error and empirical error is determined by the sample size and model capacity. The gap becomes smaller with increasing of sample size, while model design tries to minimize the expected error by defining a function space to minimize the empirical error and control the model capacity. Starting from the success of AlexNet~\cite{NIPS2012_4824} on ILSVRC-2012 dataset~\cite{ImageNet,ILSVRC}, years of effort has been devoted to model designing, and a series of improved DCNNs such as ZFNet~\cite{zeiler2014visualizing}, VGGNet~\cite{simonyan2014very}, GoogLeNet~\cite{szegedy2015going}, ResNet~\cite{he2015deep} are proposed. There are also many efforts to create new datasets for new recognition tasks~\cite{krishna2016visual, COCO, xu2016msr, zhou2014learning}. However, there is little effort to increase an existing dataset to make the empirical error closer to the expected error, mainly for two reasons, one is the labeling cost scales linearly with the size of dataset, the other is unfair comparison issue due more human labeling is used to achieve better results.

In this work, we attempt to automatically augment\footnote{This is different with the common practice of data augmentation for DCNN training, which randomly cropping training samples from an image to avoid overfitting and achieve translation/scale invariance.} an existing dataset from the Web with a pre-trained DCNN on the existing dataset.

Web hosts massive images with rich contextual information and the volume is keeping growing fast, which made many applications possible such as image search engines. Web is also the basic source of many datasets which are scraped from search engines w/o further human labeling, e.g., ImageNet~\cite{ImageNet}, Places~\cite{zhou2014learning}, 80M tiny images~\cite{torralba200880}, CIFAR-10/100~\cite{krizhevsky2009learning} etc. An image on a Web page often comes with rich contextual information edited by Web authors, e.g., Alt text that conveys the same essential information can be used for displaying to replace the associated image in a pure text-based browser, page title describes what is the whole web page about, and surrounding texts around the image which are related to the image content in some manner. Since contextual information is not purposely edited to annotate image content, it is also quite noisy.

DCNNs trained on large scale datasets have achieved superior performance, which inspires us to investigate the possibility to use DCNN replace human to do image labeling task. In our early study, we found that DCNN trained on ImageNet performs much worse on Web images due both images and categories are not following the same distribution as the training set, and results in many false positives for each category. The problem can be alleviated by setting high thresholds for the prediction score, however, these images can provide limited additional information to improve the pre-trained DCNN since the DCNN is already quite confident on these images.

DCNN extracts image's visual information while Web provides image's contextual information, which are complementary and can jointly provide additional information to an existing dataset. Noise of contextual information can be removed by the DCNN using visual information, while rich contextual information helps lower the threshold for the prediction score of a DCNN that required to achieve high prediction accuracy. Together, we can augment an existing dataset in a scalable, accurate and informative way. Specifically, we automatically augment  ILSVRC-2012 with additional 12.5 million images from the Web. By training the same DCNN on the augmented dataset without human labeled images, significant performance gains are observed, which demonstrates a well-trained DCNN can improve itself by self labeling more images from Web. Another encouraging experimental result is that we can even boost the performance of ResNet-50 on ILSCRV-2012 validation set from 74.55\% to 77.35\% by using our augmented dataset which is labeled a lower performance AlexNet. We release the dataset and models\footnote{The dataset and models can be found at \url{https://auto-da.github.io/}} to facilitate the research on learning based object recognition. 

The rest of this paper proceeds as follows: After an overview of related work in Section~\ref{related_work}. We introduce our proposed method which could automatically augment dataset according to Web labeling and DCNN labeling in Section~{method}. We evaluate the quality of augmented datasets in Section~\ref{experiment}, and conclude with discussion in Section~\ref{conclusion}.

\section{Related Work}\label{related_work}
Datasets are the basic inputs for statistical learning algorithms to train learning models, and significant efforts have been made to construct datasets for various recognition tasks. In this section, we discuss related efforts according to the amount of human labeling used during dataset construction.
%Recently, there have been increasing research interests in automatically constructing image datasets by exploiting web images from the Internet. 
\subsection {No human labeling}
Some datasets are directly collected from image search engines or social networks without human labeling. TinyImage~\cite{torralba200880} contains 80 million $32\times32$ low resolution images, collected from image search engines by using words in WordNet as queries. YFCC100M~\cite{thomee2016yfcc100m} is another large database of approximately 100 million images associated with metadata collected from Flickr. Krause et al.~\cite{krause2015unreasonable} try only using Web images to fine-tune DCNN pre-trained on ILSVRC-2012 for fine-grained classification, and get even higher accuracies than using fine-grained benchmark datasets, which is expected due existing fine-grained benchmark datasets are quite small. Phong et al.~\cite{NoisyWebImages} collect 3.14 million Web images from Bing and Flickr for the same 1000 categories of ILSVRC-2012.%, DCNN trained on this dataset performs much worse than ILSVRC-2012 which reflects the nature of Web images, which is noisy and high biased.
Massouh et al.~\cite{massouh2017learning} also proposed a framework to collect images from Web and using a visual and natural language concept expansion strategy to improve the visual variability of constructed dataset. Recently, Li et al.~\cite{webvison} also constructed a dataset by directly querying images from Flickr and Google Images Search. However, DCNN trained on all of these automatically constructed datasets perform much worse than human labeled dataset when testing on ILSVRC-2012, which reflects the noisy and high-biased nature of Web images.

\subsection {Fully human labeling} 
Each image is manually labeled by one or multiple users to ensure high accuracy. Due the high labeling cost, datasets constructed by fully labeling are often with small size, some typical exemplar datasets are Caltech101/256~\cite{fei2007learning, griffin2007caltech}, Pascal VOC~\cite{voc2007} and several ones for fine-grained object recognition~\cite{khosla2011novel,maji2013fine,CUB200-2011}. These datasets are widely used for shallow model learning, while are not large enough to train a DCNN from scratch. Though challenging, million scale datasets have been constructed, such as ImageNet~\cite{ImageNet} for object recognition and Places~\cite{zhou2014learning} for scene recognition. With ImageNet, DCNN first proves its success and improves most object recognition tasks by the learned feature extractors~\cite{NIPS2012_4824}. However, the high labeling cost limits both the number of images can be labeled for each category and the number of categories can be labeled. 

\subsection {Partially human labeling}
To alleviate human labeling cost and use limited budget in more effective ways, several active learning based approaches are proposed to only label images that are considered as informative for a model. Collins et al.~\cite{collins2008towards} propose to iteratively do image labeling and model training, where some randomly selected images are first labeled as seed training set to train an initial model, then the model is applied to a set of unlabeled images, to select a subset of images which the model is mostly uncertain for human labeling. The process is iterated until the classification accuracy converges or the budget is run out. Krause et al.~\cite{krause2015unreasonable} present 
a similar scheme for fine-grained object recognition by using DCNN as model. Since informative images are selected based on some specific model, human involvement is always required for newly designed models. 

To decouple human labeling from model training, Tong et al.~\cite{xiao2015learning} propose to train DCNN for clothing classification with both clean dataset manually labeled by annotators and millions images with noisy labels provided by sellers from online shopping websites. Though noisy, the accuracy of images from online shopping websites ($\sim 62$\%~\cite{xiao2015learning}) is much higher than general Web images ($\sim 10\%$~\cite{torralba200880}). Sukhbaatar et al.~\cite{sukhbaatar2014training} try to train DCNN with 0.3M clean ILSVRC-2012 training images and 0.9M noisy Web images, and show marginal improvement with a noise layer to model noise, but still with much higher error rate than DCNN directly trained on 1.2M ILSVRC-2012 training images.

Since the accuracy of Web images is relatively low, the number of Web images needs to be orders of magnitude larger than existing datasets to contain enough clean images. Thus, we aims to use as more Web images as possible, till $30^{th}$ July 2017, we have used 186.4 million Web images as candidate images to augment several labeled image datasets. These augmented image datasets achieve high performance on objects recognition tasks than human-labeled datasets with significantly more training images. To the best of our knowledge, this is the first work that uses DCNN to label Web images and demonstrates a well-trained DCNN can automatically improve itself by ``surfing'' the Web.

\section{Automatic Dataset Augmentation}\label{method}

Starting from a human labeled image dataset $\mathcal{D}$, we are targeting at augmenting it to a much larger dataset $\mathcal{D} \cup \mathcal{E}$, where $\mathcal{E}$ is automatically labeled from Web images by a DCNN trained on $\mathcal{D}$. Labeling images is an intelligent process which requires sufficient intelligence and knowledge. In this section, we will first investigate two separated labeling methods by DCNN and Web, respectively, then present our method which labels image by both Web and DCNN. If no special mention is made, AlexNet designed by Krizhevsky \etal~\cite{NIPS2012_4824} will be used as the basic DCNN considering it is with relatively low computational cost for large scale experiments.
%enlarging this image dataset $E$ from large scale unlabeled user click dataset $U$ automatically in a scalable way while ensure both accuracy and diversity.

%Considering the diversity and coverage are still concerns as most search engine only returning limited amount of images, and most of them are click attractive and representative images, we directly use billions of Internet images as unlabeled candidate image dataset rather than the search engine returned images. %Even though the Internet images lack human annotation, they often have some kind of noisy label gleaned the image. Some noisy data can be easily reduced, while some other noisy data is hard to be found and removed unless some extra informations are used.

\subsection{Labeling By DCNN}
% such as user click information and semantic relevance measures are used. Moreover, we increase the diversity of the constructed dataset by expanding the relevance queries for each category.
DCNNs have achieved remarkable prediction accuracy on validation set and testing set of ILSVRC-2012~\cite{ILSVRC} by end-to-end learning on the training set, which inspires us to use DCNN replace human to do the image labeling task. Given a DCNN trained on the labeled dataset $\mathcal{D}$, which maps an image $I$ to a set of confidence scores $f_c(I)$ for each pre-defined category $c\in\{1,...,C\}$. Then, using the DCNN to do labeling is intuitive, a new image $I$ can be labeled as an instance of category $c$ if $I$ has a confidence score on $c$ exceeds some threshold $\alpha$, i.e.,
\begin{equation}
f_c(I) \geq \alpha,
\end{equation}
Then an augmented dataset $\mathcal{E}_V$ can be labeled by applying the DCNN on a large unlabeled image set $\mathcal{U}$, i.e.,
\begin{equation}
\mathcal{E}_V=\big\{\langle I,c \rangle:f_c(I)\geq \alpha, I\in\mathcal{U}, c\in\{1,...,C\}\big\}.\label{v}
\end{equation}
The labeling process is fully automatic which only requires feedforward calculation on a unlabeled image set. We investigate this method by using the DCNN learned from ILSVRC-2012 training set to label an unlabeled candidate image set which randomly collected from Web. By analyzing the labeling results, we find several properties of labeling by DCNN.

\begin{figure}[ht]
	\centering
	\includegraphics[width=0.48\textwidth,page=23]{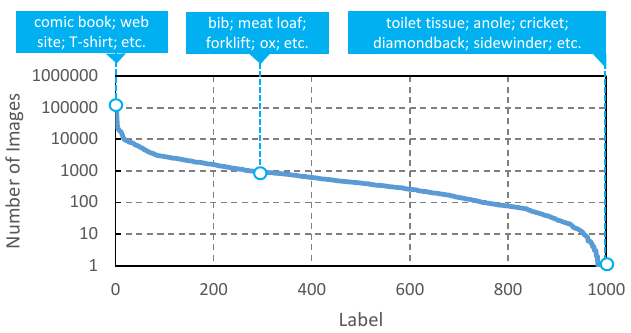}
	\caption{The label frequencies of dataset labeled by DCNN with confidence threshold 0.9. The distribution of labels across the images is highly uneven with some labels have more than ten thousand images, while others have less than one hundred.}\label{heavy_tail}
\end{figure}

\textbf{Unbalance} Figure~\ref{heavy_tail} shows the number of images labeled for each category using a relatively high threshold $\alpha=0.9$. The number of images of different categories is extremely unbalanced, where ``web site'' has more than 100,000 images, while ``toilet tissue'', ``American chameleon, anole'' have no images. The unbalance is caused by the unbalanced nature that web images since images of some categories are inherently more popular than others. To label enough images for each category, the only way is to predict more images where most computations are spent on images of popular categories.

\begin{figure}[ht]
	\centering
	\includegraphics[width=0.47\textwidth,page=25]{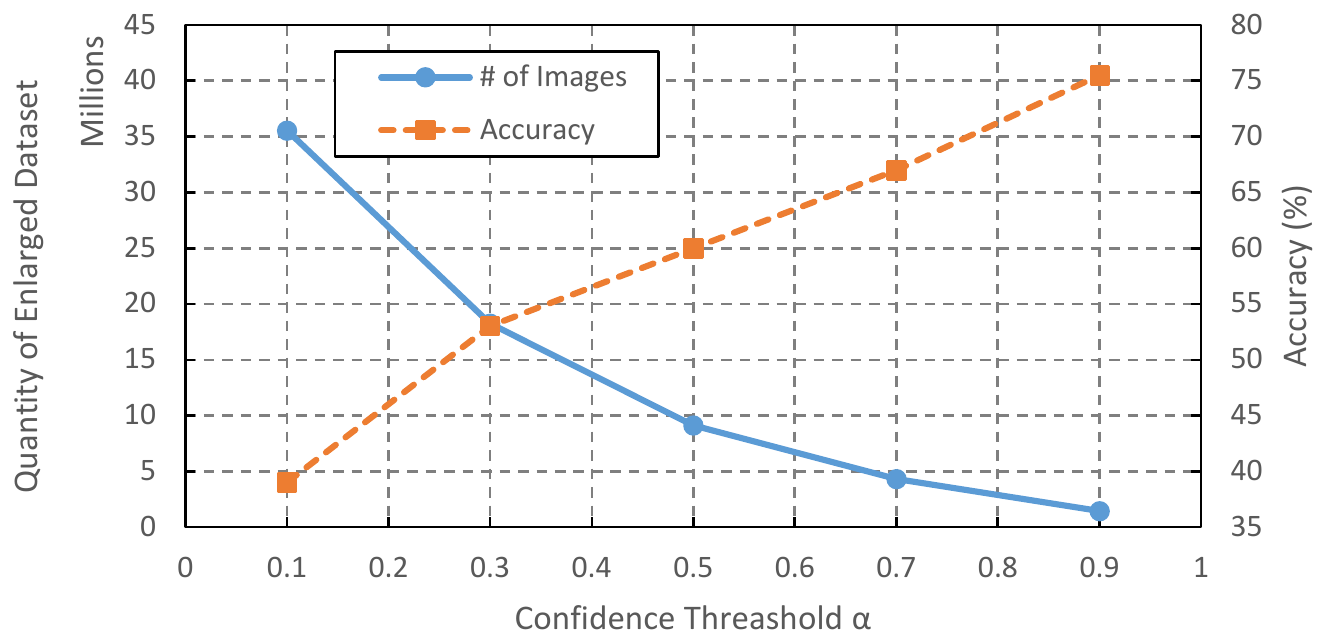}
	\caption{The distributions of quantity and accuracy of dataset $\mathcal{E}_V$ across confidence threshold $\alpha$.}\label{SLC}
\end{figure}

\begin{figure}[ht]
	\centering
	\includegraphics[width=0.47\textwidth,page=15]{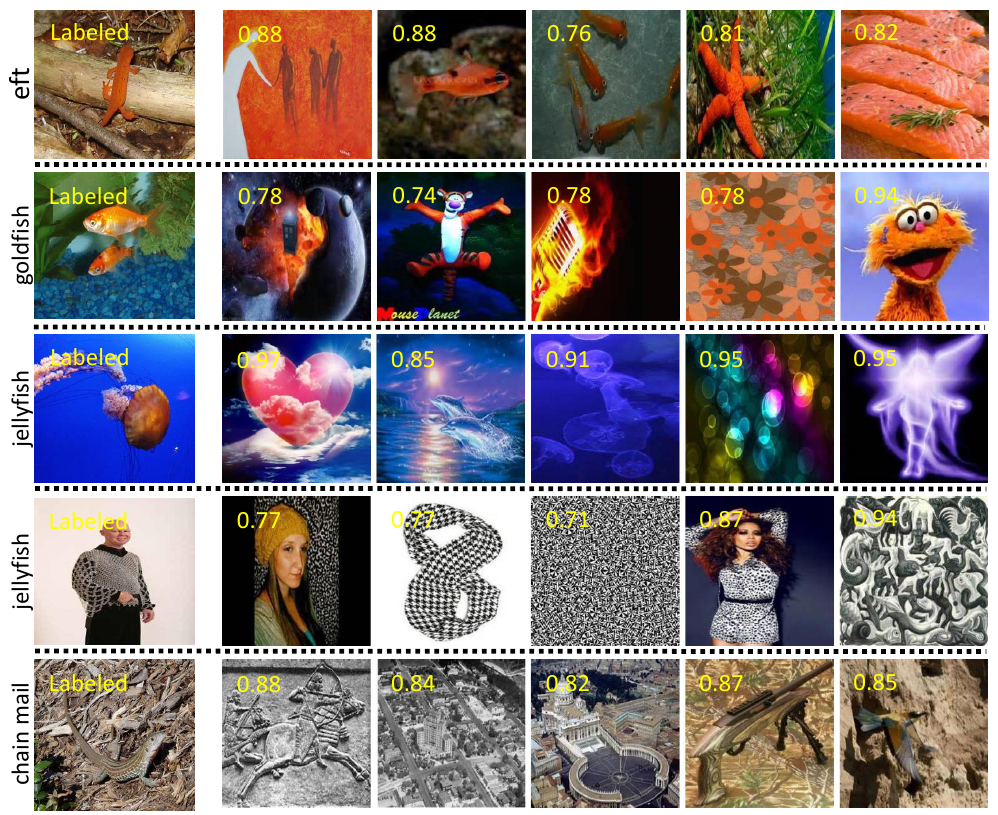}
	\caption{Noisy images which predicted as one of the categories with high confidence in labeled images dataset by DCNN. The first column in this figure shows an example image from labeled image for each category. The other columns show noisy images with high-confidence DCNN predictions for the categories in different row respectively. The confidence scores are shown on each noisy image.}\label{noisy_image}
\end{figure}

\textbf{Low Accuracy} Figure~\ref{SLC} shows the quantity and accuracy of automatically labeled datasets by setting different thresholds $\alpha$, where accuracy is estimated by manually inspecting randomly sampled images (10 random images per category) from 100 categories in constructed dataset. As expected, higher threshold will result smaller dataset with higher accuracy. However, even with the relatively high threshold $0.9$, the accuracy $75.5\%$ is still much lower than the accuracy $99.7\%$ achieved by human labeler on ImageNet~\cite{ImageNet}. %We find that the performance of classification decreases when we try to add high-confidence instants from $U$ to $E$. We find that the noise in enlarged dataset can be considered as two different types.
%\textbf{Unrecognizable Noise}
%It main due to the noise in the enlarged dataset. We consider two types of label noise:
%\subsubsection{Noise Analysis}\label{noise_analysis}
%The trained DNN usually have good performance on classifying the categories which have labeled images for training.
Figure~\ref{noisy_image} shows some false positive images that are incorrectly labeled for each category, where most images are out of the 1000 categories used for training but visually similar to the category in some aspects. The result also shows that the DCNN is still hard to generalize to a testing set with many out-of-class images.

\begin{figure*}[ht]
	\centering
	\includegraphics[width=1.0\textwidth,page=1]{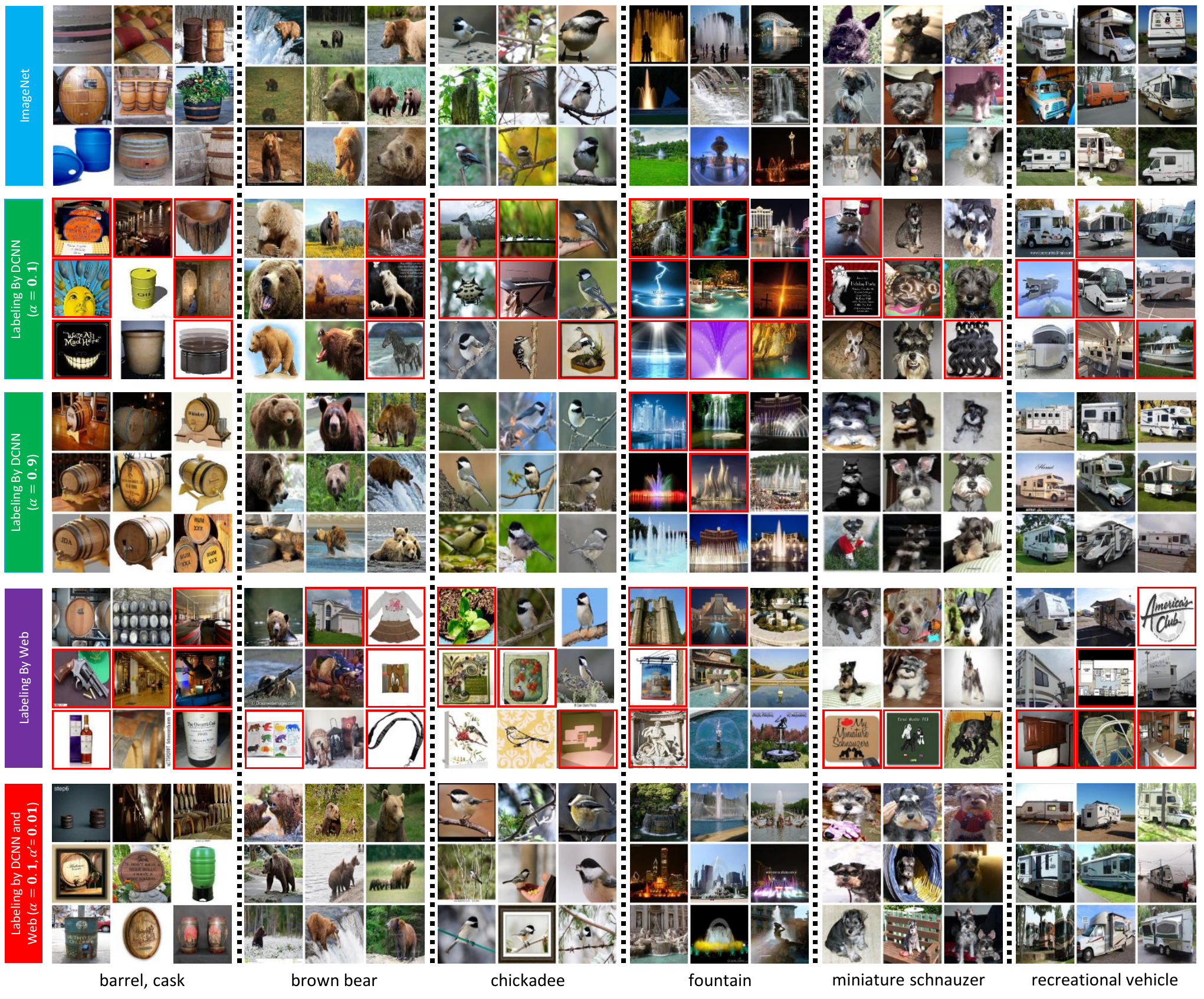}
	\caption{Snapshots of human labeled dataset ImageNet and four automatically constructed datasets on 6 randomly sampled categories in ILSVRC-2012: the first row is from the ImageNet; the second and third row are from the dataset labeled by DCNN with confidence threshold $\alpha = 0.1$ and $\alpha = 0.9$ respectively; the fourth row is from the dataset labeled by Web; the last last row is from the dataset labeled jointly by DCNN and Web with confidence threshold $\alpha = 0.1, \alpha'=0.01$. For each category, 9 randomly sampled images are presented. Images marked with red boxes are noisy images.}\label{data_view}
\end{figure*}

\textbf{Less Informative} Though higher accuracy can be obtained by keeping increasing the threshold, this will cause two problems. One is the number of images can be collected will be reduced for a fixed unlabeled dataset, and the unlabeled dataset needs to be even larger to collect enough images. The other problem is even worse, images labeled by high confidence scores are iconic samples and with high similarity with images in existing training set as showed in the third row of Figure~\ref{data_view}, which can bring little new supervisory information to the existing training set.

%To further remove the noise data which can be hard filtered out by visual restriction and increase the diversity of enlarged dataset, the extra semantic restriction is necessary.% we propose a framework to use multiple textual metadata in the process of constructing dataset. The follow subsections describe the details of our method.

%After recognizing the unlabeled images by using the DNN trained on labeled image dataset, we find that not only some unrecognizable image are predicted with high confidence, but also lots of images that are not of any category in the labeled dataset recognized by DNN as one of category in labeled dataset with high confidence as Figure~\ref{noisy_image} shown.

\subsection{Labeling By Web}
\begin{figure}[ht]
	\centering
	\includegraphics[width=0.47\textwidth,page=3]{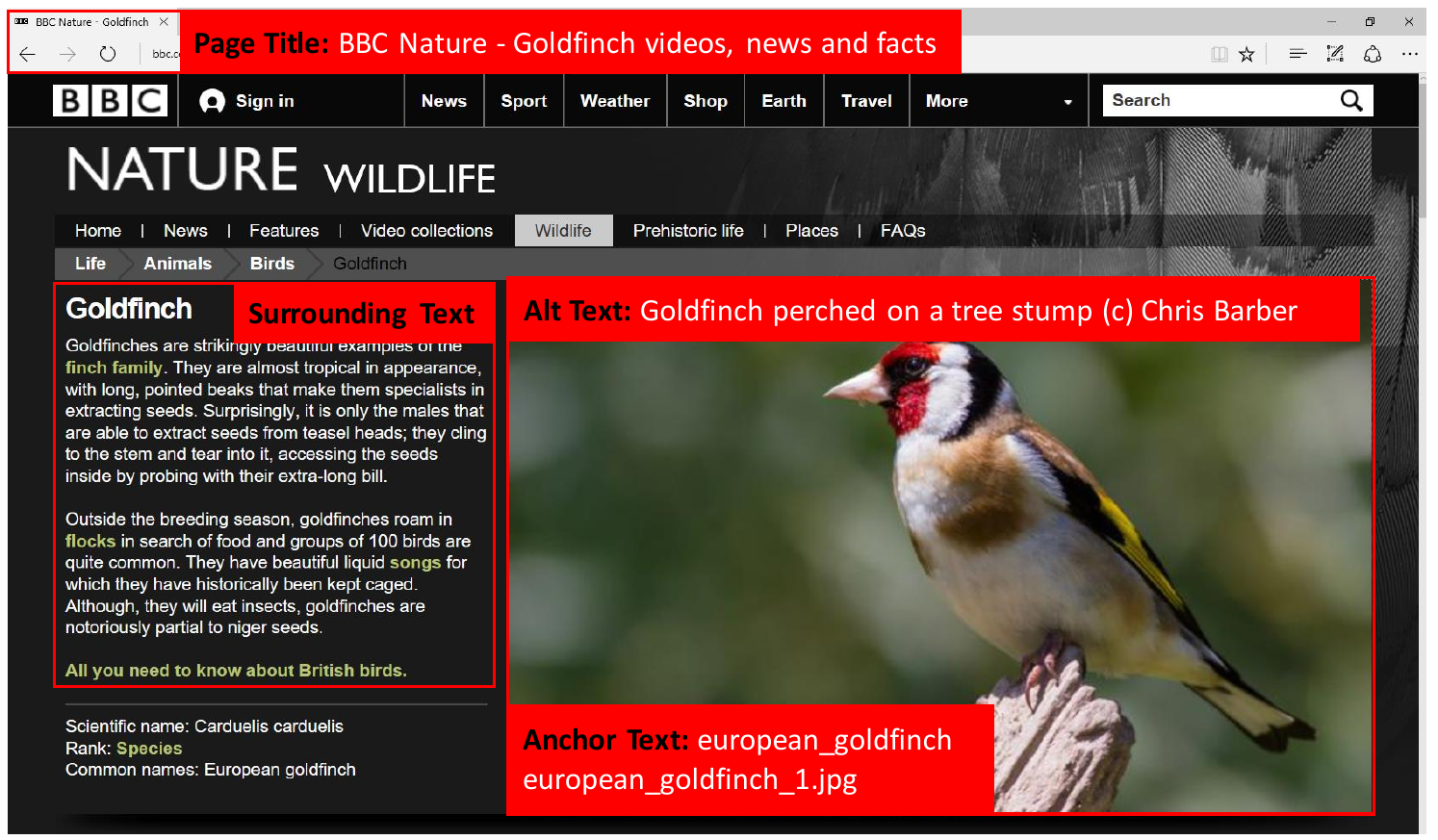}
	\caption{Illustration of the textual metadata associated with an image in a web page. The web page used in this figure is from \url{http://www.bbc.co.uk/nature/life/European_Goldfinch}.}\label{metadata_in_webpage}
\end{figure}

Web hosts trillions images with rich metadata, which provides a ``free'' way to label images since labels are already in the metadata provided by Web users. Image search engines directly leverage these metadata to index massive Web images and make them retrievable. Though image search engines provide a convenient way to collect Web images by searching words or word phrases that describe a category, they are with several limitations for dataset construction as they are optimized for human users, e.g., they typically limit the number of images retrievable for each query (in the order of a few hundred to a thousand), the retrieved images are often iconic, presenting a single, centered object with a simple background, which is not representative of natural conditions. Thus, we directly resort to raw images with textual metadata from the Web as our source data. Specifically, four textual fields are collected for each image, including
\begin{itemize}
  \item \emph{Anchor text} $T^1$ is the visible, clickable text in a hyperlink linked to an images, which usually gives the user relevant description about the content of the linked image.
  \item \emph{Alt text} $T^2$ is shown when an image cannot be displayed to a reader. Thus, it can be seen as a textual counterpart to the visual content of an image. %Many are not provided.
  \item \emph{Page title} $T^3$ is an important field for the page author to state what the main content of the webpage is about.
  \item \emph{Surrounding text} $T^4$ consists of the text paragraphs around an image in a webpage. The surrounding text is in many cases semantically related to the image content. However, since the surrounding text can also contain information that is uncorrelated to the image, this field as a contextual information source can be much noisy.
\end{itemize}
Then a data item from Web can be denoted by $\langle I,T^1,T^2,T^3,T^4 \rangle$. Figure~\ref{metadata_in_webpage} shows a web image and its four types of textual metedata, where rich information about is embedded in metadata for the image.

%Since textual data is much smaller than visual data, we crawled xxx webpages and stored their textual metadata. Give a set of categories described by words or word phrases, we use simple string match to find candidate images for each category. %The number of images for different categories are extremely unbalance, since some categories are quite common while some are rare. To balance, we set a upbound for the number of each category.

Given a web image dataset denoted by $\mathcal{W}=\{ \langle I_i,T^1_i,T^2_i,T^3_i,T^4_i \rangle\}_{i=1}^{|\mathcal{W}|}$, labeling by Web can be directly carried out through string match. Let each category $c$ be represented by a set of word phrases from its WordNet synonyms~\cite{WordNet} and relevant descriptions in 12 different languages (including AR, ZH, EN, FR, DE, EL, HE, HI, IT, JA, RU, ES) from BableNet~\cite{NavigliPonzetto:12aij}, denoted by $\mathcal{S}_c =\{s_j\}_{j=1}^{|\mathcal{S}_c|}$. An image $I_i$ is labeled as an instance of category $c$ if at least one textual field contain at least one element in $\mathcal{S}_c$, i.e.,
\begin{equation}
\delta_{i}^c = \left\{
\begin{array}{ll}
1&: s_j \subseteq T_i^k, \exists s_j \in \mathcal{S}_c, \exists k\in\{1,...,4\}\\
0&: otherwise
\end{array}
\right.
\end{equation}
Then an augmented dataset $\mathcal{E}_T$ can be labeled by Web data $\mathcal{W}$, i.e.,
\begin{equation}
\mathcal{E}_T = \big\{\langle I,c \rangle:\delta_i^c=1, i\in \{1,...,|\mathcal{W}|\}, c\in\{1,...,C\}\big\}
\end{equation}
The labeling process is also fully automatic and very fast after $|\mathcal{W}|$ has been collected.

By the method, we collect a dataset with 186.4 million images for the 1000 categories from ILSVRC-2012 dataset. Here, we summarize several properties observed from the dataset.

\begin{figure}[ht]
	\centering
\includegraphics[width=0.45\textwidth,page=13]{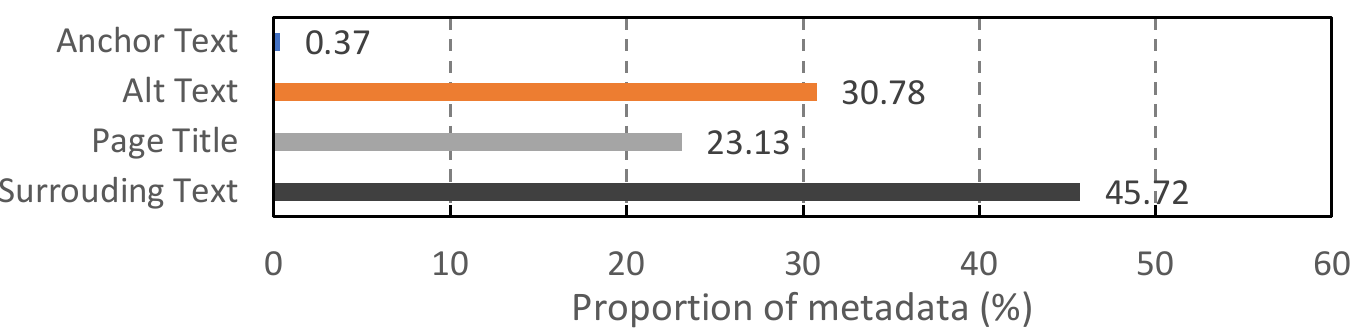}
\caption{The proportion of images collected according to different fields of textual metadata.}\label{textual_scale}
\end{figure}

Figure~\ref{textual_scale} shows percentage of images collected by each textual field, where surrounding text contributes the most since most images are with surrounding texts and typically contains more words than other fields, while the number of images collected by anchor text is much smaller than other fields since anchor texts are typically very short and often not provided by web authors.

\begin{figure}[ht]
	\centering
	\includegraphics[width=0.4\textwidth,page=2]{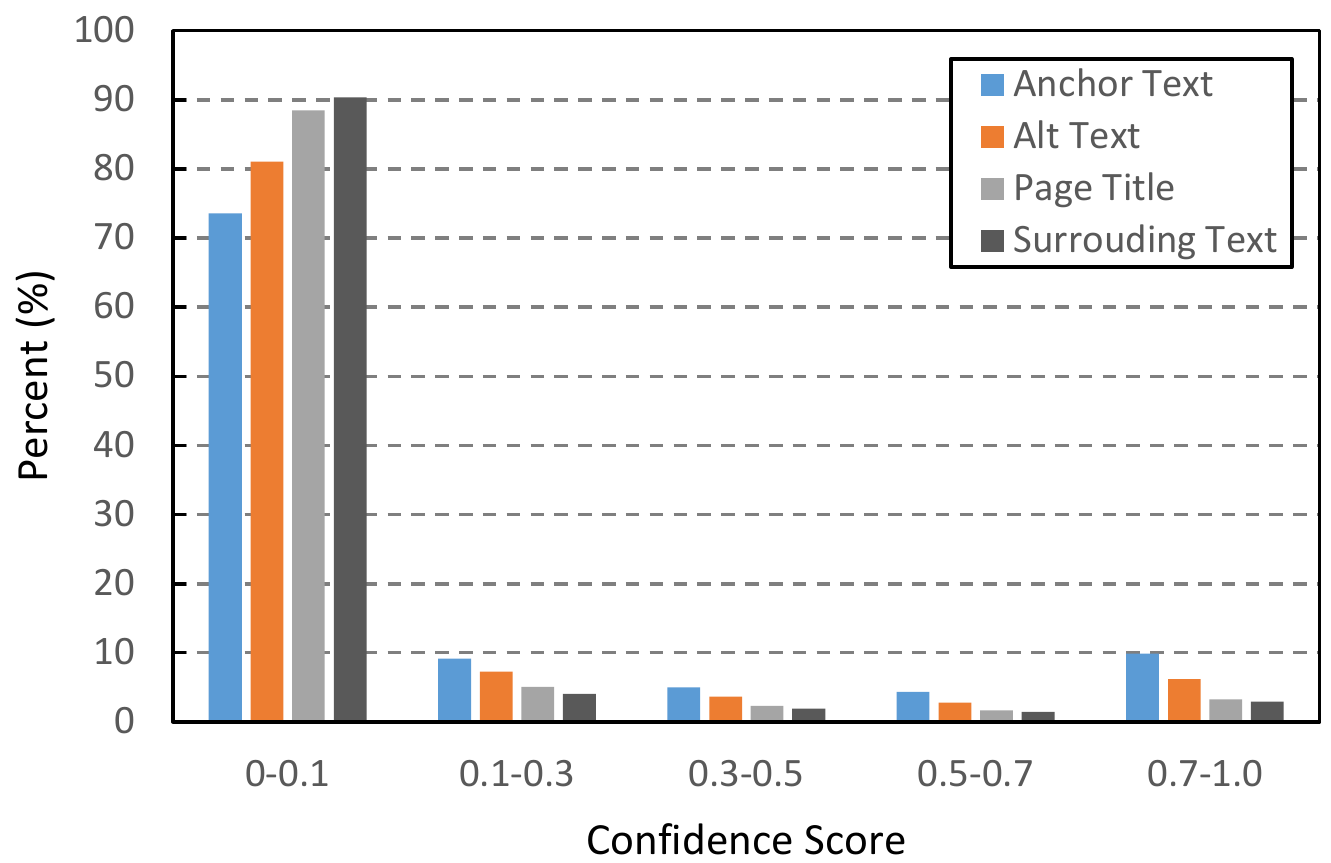}
	\caption{The distributions of confidence score across the percent of images constructed according to different kinds of contextual information.}\label{confidence_distribution}
\end{figure}

Besides the quantity, we also check the quality of the collected dataset. To avoid manually checking, we use the DCNN to calculate the confidence score of the labeled category of each image in $\mathcal{E}_T$, and large confidence score means large probability of the labeled image to be correct. Figure~\ref{confidence_distribution} shows the distribution of confidence scores by different textual fields, where images collected by anchor text and alt text are with larger proportion of high confidence scores, which also means these two fields are more reliable than the others. The conclusion is also consistent with experiences of using textual features for image search engines~\footnote{https://support.google.com/webmasters/answer/114016?hl=en}.

However, as expected, images collected from Web are very noisy, where 82.8\% images are with confidence scores lower than 0.05. After analyzing the noisy images, we find that most noisy images are introduced by ambiguities in textual metadata. A typical example is a category named by ``jay'' which is supposed to be a bird, lots of images about human are collected since ``jay'' is often used as human name. Though these noisy images are hard to remove by only using textual information, they are easy to remove by visual information since images of different senses of a name are typically visually distinguishable.

\subsection{Labeling by Web and DCNN}
\begin{figure}[ht]
	\centering
	\includegraphics[width=0.45\textwidth,page=30]{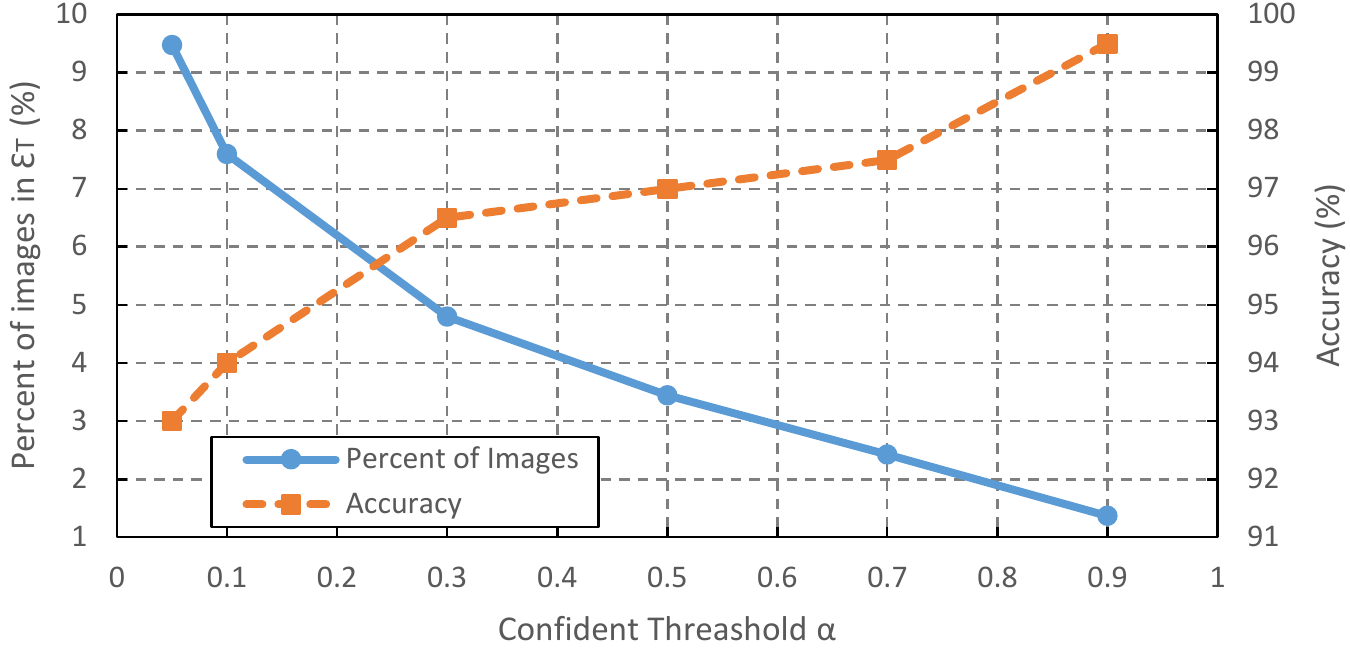}
	\caption{The distributions of quantity and accuracy of dataset $\mathcal{E}_{VT_{web}}$ across confidence threshold $\alpha$ after applying visual restriction to candidate dataset $\mathcal{E}_T$.}\label{visual_restriction_results}
\end{figure}

Since both visually labeling by DCNN and contextually labeling by the Web have their own limitations, here we combine them together to improve the labeling by leveraging their complementarity. Learned from the above experience that labeling by DCNN is more computational cost and tend to spend too much time on popular categories, thus we first use the Web to label a dataset $\mathcal{E}_T$  in a relatively balanced way, then use DCNN to go through the textually labeled dataset $\mathcal{E}_T$. Together, a dataset can be labeled by Web and DCNN via,
\begin{equation}
\mathcal{E}_{VT_{web}}=\big\{\langle I,c \rangle:f_c(I)\geq \alpha, \langle I,c \rangle \in\mathcal{\mathcal{E}}_T \big\} \label{vt}
\end{equation}
where $\mathcal{E}_{VT_{web}}$ is a filtered subset of $\mathcal{E}_{T}$ where lots noisy images are filtered out by DCNN. Different from labeling by DCNN in Eq.~\ref{v}, the contextual labeling can filter out the majority of out-of-class noisy images, and the used $\mathcal{E}_{T}$ is with much higher signal-noise ratio than $\mathcal{U}$, which allows us to use lower threshold $\alpha$ to label more informative images. Figure~\ref{visual_restriction_results} shows the quantity and accuracy curve with respect to confidence threshold $\alpha$ on images labeled by the Web, it is encouraging that much higher accuracy is achieved even with very low confidence threshold, e.g., 94\% accuracy is achieved when the threshold $\alpha$ is set to 0.1. 

%%考虑到ET的accuracy还是非常的低，如果我们可以进一步减少ET的noise的话，那么就可以在保证accuracy的情况进一步降低DCNN的threshold，从而进一步扩大dataset的scale同时提升diversity。
Since the accuracy of $\mathcal{E}_{T}$ is still relatively low by simply using string match, which limits us to set lower confidence threshold to bring in more diverse and difficult images while keep high accuracy. Thus, we are motivated to further decrease the noise in $\mathcal{E}_{T}$. Since image $I_i$, text $T_i$, metadata type $t_i$ and image URL domain $d_i$ are coupled together as a single data item in our dataset, labels assigned to images by DCNN are also assigned to metadata, then we construct an automatically labeled textual dataset, i.e.,
\begin{equation}
\begin{aligned}
\mathcal{T}^{+} &= \big\{\langle T_i, t_i, d_i, y_i = c_i \rangle: \langle I_i, c_i \rangle \in \mathcal{E}_{VT_{web}} \big\}\\
\mathcal{T}^{-} &= \big\{\langle T_i, t_i, d_i, y_i = C + 1 \rangle: I_i \in \mathcal{N}_{VT_{web}} \big\}\\
\end{aligned}
\end{equation} 
where $\mathcal{N}_{VT_{web}} = \big\{\langle I,c \rangle:f_c(I)< \beta, \langle I,c \rangle \in\mathcal{\mathcal{E}}_T, \beta \ll \alpha \big\}$ collects noisy images for each category by string match.
Inspired by previous work on sentence classification~\cite{joulin2016bag}, we train a two-layer fully connected network to categorize textual metadata at semantic level. The input to the network is the one hot representation of metadata type $t_i$, image URL domain $d_i$ and bigrams in $T_i$. As Figure~\ref{confidence_distribution} shown, the metadata type $t_i$ is a helpful prior for this text classification task. Meanwhile, we also found that there are some special websites on which vast majority of images are relevant to a some category, e.g. ``farnhamanglingsociety.com'' is a website about fishing and lots of images about tench can be found on this website. The first layer of the network generates embedding representation for inputs with weight matrix $E$, and the second layer classifies into categories based on the representation with weight matrix $W$ using softmax regression, 
\begin{equation}
\begin{aligned}
p(y_i=c \mid  T_i, t_i, d_i) &= \frac{e^{f(y_i=c \mid T_i, t_i, d_i)}}{\sum_{k=1}^{C+1}e^{f(y=k \mid T_i, t_i, d_i)}}, \\
f(y=k \mid T_i, t_i, d_i) &= \left ( W_k \frac{\sum_{s_j\subseteq T_i}\!\!E\cdot s_j \!\!+\!\!E\cdot t_i \!\!+\!\! E\cdot d_i}{\left | T_i \right | + 2} \right )
\end{aligned}
\end{equation}
The model is trained by minimizing
\begin{equation}
-\frac{1}{N} \sum_{i=1}^{N}\sum_{k=1}^{C+1}1\{y_i=k\}\log p(y_i=k\mid T_i, t_i, d_i)
\end{equation}
where $N = \left | \mathcal{E}_{VT_{web}}  \right | + \left | \mathcal{N}_{VT_{web}}  \right |$. We train this model by using stochastic gradient descent and a linear decaying learning rate. As a result, a new dataset $\mathcal{E}_{VT_{web^+}}$ labeled by our text classification model can be constructed:
\begin{equation}
\begin{aligned}
\mathcal{E}_{T_{web^+}} = \big\{\langle I,c \rangle: & p(y=c\mid T_i, t_i, d_i) > 0.5, \\
& i\in \{1,...,|\mathcal{W}|\}, c\in\{1,...,C\} \big\}
\end{aligned}
\end{equation}
The experimental results show that the accuracy of images set $\mathcal{E}_{T_{web^+}}$ is nearly 71.5\%, which is significantly higher than $\mathcal{E}_{T}$ whose accuracy is nearly 21.3\%.%  constructed from Web can be improved from 21.3\% to 68.9\% by using our proposed text classification instead of simple string matching.
Naturally, a new dataset with jointly constrained by DCNN and text classification model can be constructed:
\begin{equation}
\mathcal{E}_{VT_{web^+}} =\big\{\langle I,c \rangle:f_c(I)\geq \alpha', \langle I,c \rangle \in\mathcal{E}_{T_{web^+}} \big\} \label{vt}
\end{equation}
where $\alpha' < \alpha$. The high-performance text classification model makes it possible to decrease the visual threshold from $\alpha$ to $\alpha'$, and to mine  a more diverse and larger scale dataset without accuracy dropping, e.g. 93.8\% accuracy is achieved when $\alpha' = 0.01$. Finally, we get a dataset labeled by Web and DCNN jointly,
\begin{equation}
\mathcal{E}_{VT} = \mathcal{E}_{VT_{web}} \bigcup \mathcal{E}_{VT_{web^+}}
\end{equation}

 % the accuracy of our proposed text classification model can archive 68.9\% on test set, which is much higher than accuracy of string matching 21.3\%.

%% 如果可以区分出正负样本集合，那就可以在句子语义级别而不是简单的文本matching上对image进行label。从而进一步提升

Figure~\ref{data_view} shows snapshots of human labeled dataset ImageNet and four automatically constructed datasets by different methods. Compare to the dataset labeled only by DCNN or the Web, the dataset constructed based on semantic and visual jointly restriction could have higher accuracy and diversity.

\section{Experimental Results}\label{experiment}
In our experiments, for a given category set with labeled images, we first train a DCNN that will be used for labeling and as the baseline for comparing.
%a deep convolutional neural network model for classification. Then label the candidate images collected from web, all images whose confidence score is larger than visual restriction threshold $\alpha=0.1$ are kept.
%Considering inference procedure of the state-of-the-art complex image recognition models such as ResNet~\cite{he2015deep} which requires a significant amount of computing resources, especially for predicting millions of images, and the quality of constructed dataset less dependent on the performance of recognition model as we described in section~\ref{jointly_restriction}, we used the simple AlexNet architecture~\cite{NIPS2012_4824} as classifier for all of the experiments in this paper. It worthy note that the visual restriction threshold $\alpha$ is set as 0.1 in all of our experiments.
%In the rest of the section, we present our observations on augmented image dataset in two aspects: dataset properties and performance.
To make comprehensive analysis and comparisons, we consider four sets of categories from different domains, including dog, bird, wheeled object and structure. Human labeled datasets for the four domains are subsets of ILSVRC-2012 training set. Four DCNNs trained on each human labeled dataset are used to label a Web labeled dataset $\mathcal{E}_T$ which contains 186.4 million images. Table~\ref{scale} summarizes statistics of the human labeled datasets and the automatically labeled datasets, and our method significantly increases dataset scale in each domain.

%We plan to evaluate the quality of our automatically constructed dataset according the performance of object recognition model learned from them, but more training data requires more training time. Thus we only randomly sampled at most 10 times the amount of images in labeled dataset for each category from $\mathcal{E_{VT}}$.

%Table~\ref{scale} shows the detailed scale for these four different datasets. The image dataset constructed from unlabeled web images (no labeled dataset is included) are named with suffix ``-A''. The quantity distributions of there datasets is shown in Figure~\ref{4dataset_distributions}. It is worth noting that the datasets we described here is still increasing.

\begin{table}
	\begin{center}
		\begin{tabular}{|l|c|c|}
			\hline
			Dataset & \# of categories & \# of images \\
			\hline\hline
			$\mathcal{D}_{ImageNet}^{dog}$ & 120 & 150,473 \\
			\hline
			$\mathcal{E}_{VT}^{dog}$ & 120 & 1,287,831 \\
			\hline\hline
			$\mathcal{D}_{ImageNet}^{bird}$ & 59 & 76,541  \\
			\hline
			$\mathcal{E}_{VT}^{bird}$ & 59 & 710,059  \\
			\hline\hline
			$\mathcal{D}_{ImageNet}^{wheeled}$ & 44 & 57,059  \\
			\hline
			$\mathcal{E}_{VT}^{wheeled}$  & 44 & 607,680  \\
			\hline\hline
			$\mathcal{D}_{ImageNet}^{structure}$ & 58 & 74,400   \\
			\hline
			$\mathcal{E}_{VT}^{structure}$  & 58 & 745,940  \\
			\hline
		\end{tabular}
	\end{center}
	\caption{The scale of the labeled image datasets and augmented datasets on four domains. $\mathcal{D}_{ImageNet}$ means the human labeled images from ILSVRC-2012 training set.}\label{scale}
\end{table}

\subsection{Results on Augmented Datasets}\label{recognition_results}
In this paper, we measure the quality of augmented datasets by measuring their performance on object recognition. Single-crop top-1 accuracy on corresponding subsets of ILSVRC-2012 validation set is used as the performance metric. Table~\ref{recognition_performance} reports the results of DCNNs trained from augmented datasets and human labeled datasets. The augmented datasets (without human labeled images) give consistent improvement across all four different domains, which demonstrates that well-trained DCNNs can automatically label more useful images from Web and improve themselves further. Averaging the predictions of the two DCNNs trained on human labeled datasets and augmented datasets can further improve the performance.

\begin{table}
	\begin{center}
		\begin{tabular}{|l|c|c|c|}\hline
				\multirow{2}{*}{Domain}& \multicolumn{3}{c|}{Training Data}  \\\cline{2-4}
				& $\mathcal{D}_{ImageNet}$ &~~~$\mathcal{E}_{VT}$~~~ & $\mathcal{E}_{VT}\cup \mathcal{D}_{ImageNet}$ \\\hline
				\hline
				Dog & 65.80 & 67.56 & \textbf{70.22} \\
				\hline
				Bird & 82.00 & 86.24 & \textbf{86.41} \\
				\hline
				Wheeled & 68.95 & 72.20 & \textbf{74.59}\\
				\hline
				Structure & 66.07 & 69.90 &  \textbf{72.41} \\
				\hline
		\end{tabular}
	\end{center}
	\caption{Single-crop top-1 accuracy of DCNNs trained on human labeled datasets and augmented datasets.}\label{recognition_performance}
\end{table}

%
%\begin{table}
%	\begin{center}
%		\begin{tabular}{|l|l|c|}
%			\hline
%			Model & Training Data &  Accuracy (\%) \\
%			\hline\hline
%			1 DCNN & Dog & 65.80  \\
%			\hline
%			1 DCNN & Dog-A & 67.56  \\
%			\hline
%			2 DCNNs & Dog + Dog-A & \textbf{70.22} \\
%			\hline\hline
%			1 DCNN & Bird & 82.00 \\
%			\hline
%			1 DCNN & Bird-A & 86.24   \\
%			\hline
%			2 DCNNs & Bird + Bird-A & \textbf{86.41} \\
%			\hline\hline
%			1 DCNN & Wheeled &  68.95 \\
%			\hline
%			1 DCNN & Wheeled-A & 72.20 \\
%			\hline
%			2 DCNNs & Wheeled + Wheeled-A & \textbf{74.59} \\
%			\hline\hline
%			1 DCNN & Structure &  66.07 \\
%			\hline
%			1 DCNN & Structure-A & 69.90 \\
%			\hline
%			2 DCNNs & Structure + Structure-A & \textbf{72.41} \\
%			\hline
%		\end{tabular}
%	\end{center}
%	\caption{The performance of classifiers trained on human labeled dataset and dataset enlarged from Web.}\label{recognition_performance}
%\end{table}

\begin{figure*}[!ht]
	\centering
	\begin{minipage}[t]{0.425\linewidth}
		\centering
		\includegraphics[page=26, width=1.0\linewidth]{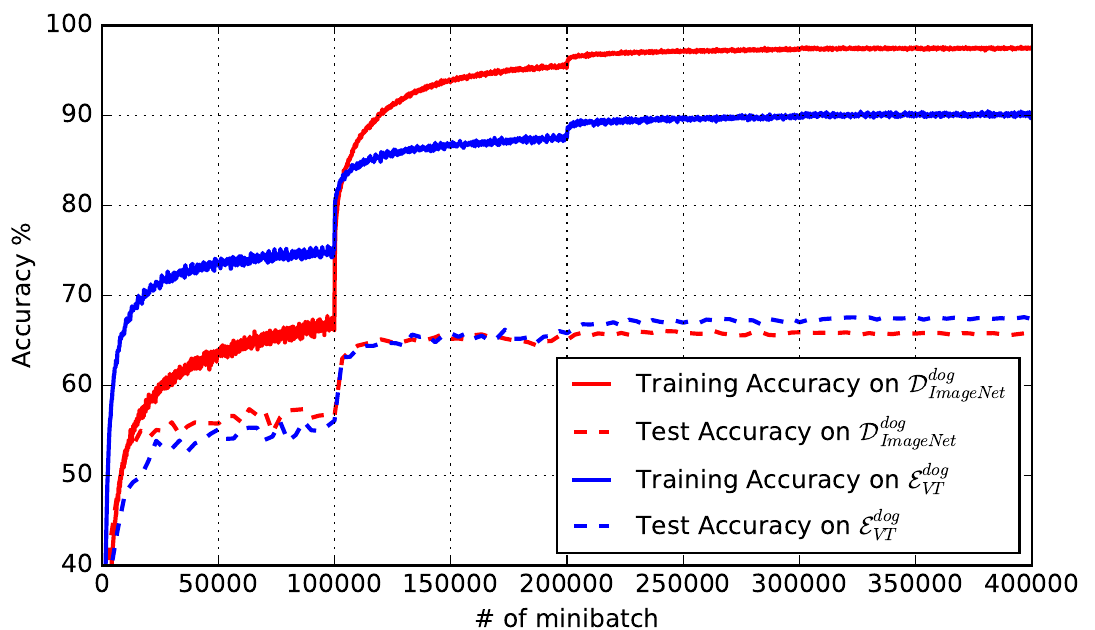}
	\end{minipage}%
	~~~~~~~~~~~~~~
	\begin{minipage}[t]{0.425\linewidth}
		\centering
		\includegraphics[page=20, width=1.0\linewidth]{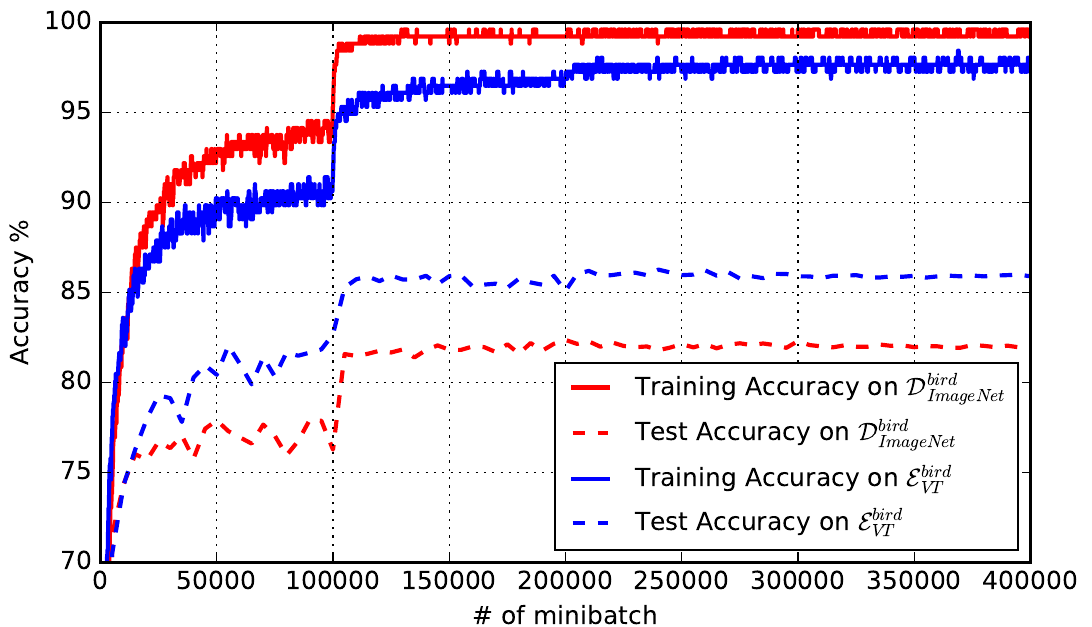}
	\end{minipage}%
	\\
	\centering
	\begin{minipage}[t]{0.425\linewidth}
		\centering
		\includegraphics[page=21, width=1.0\linewidth]{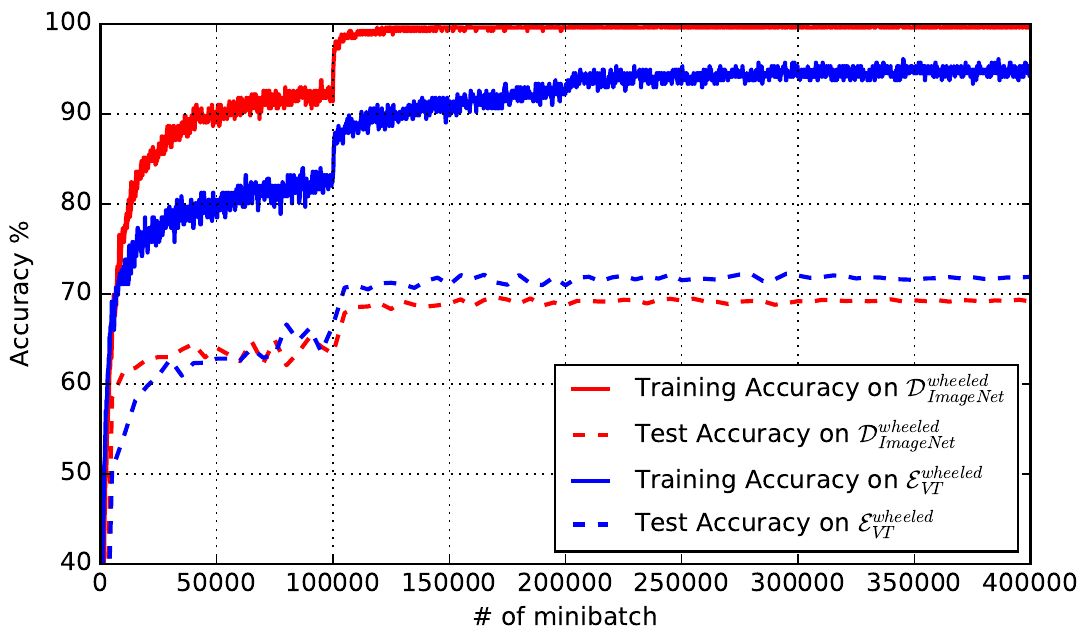}
	\end{minipage}
	~~~~~~~~~~~~~~
	\begin{minipage}[t]{0.425\linewidth}
		\centering
		\includegraphics[page=22, width=1.0\linewidth]{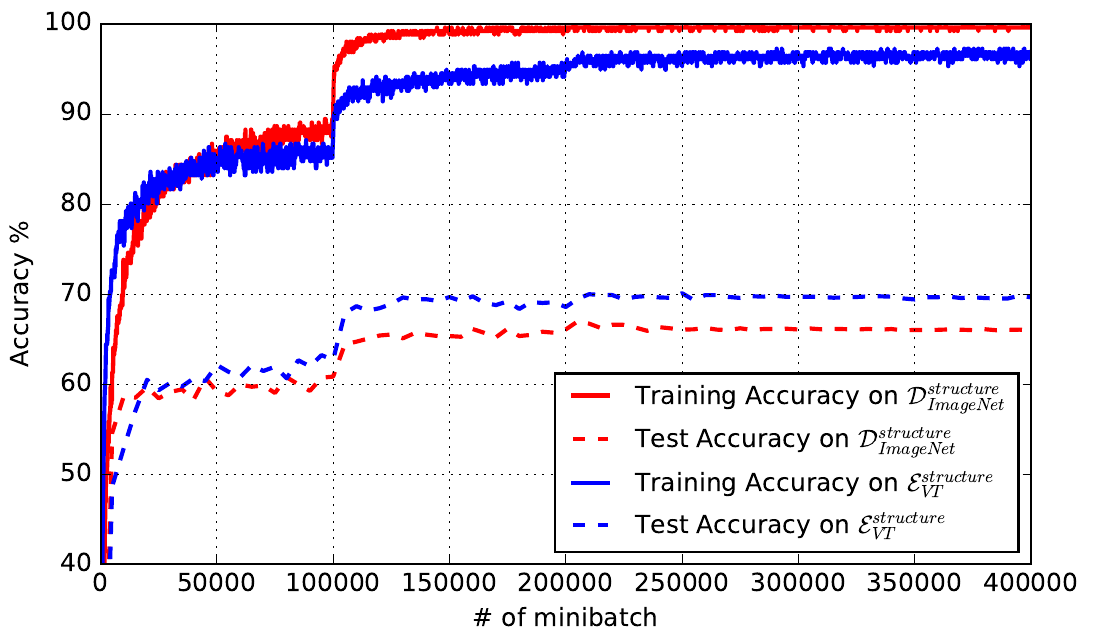}
	\end{minipage}
	\caption{Training/test curves of DCNNs learned on human labeled datasets $\mathcal{D}_{ImageNet}$ and the automatically labeled datasets $\mathcal{E}_{VT}$ from four different domains.}
	\label{overfitting}
\end{figure*}

To analysis how augmented datasets improve the performance of recognition, we compare the training/test curves of the DCNNs trained on $\mathcal{D}_{ImageNet}$ and $\mathcal{E}_{VT}$ in Figure~\ref{overfitting}. The smaller training/test gap and better test accuracy show that the significantly larger datasets help prevent overfitting for training deep models. 
%As we known, with limited human labeled training data, many of the complicated relationships between DCNN's inputs and outputs will be the result of sampling noise, and they will exist in the training set but not in real test data. This leading to overfitting. There are many methods have been developed to prevent overfitting, such as dropout and traditional data augmentation (random cropping, image flipping). However, even with using these methods in our experiments,

To analyze how Web labeling influence the quality of constructed dataset, we compare the performance of DCNNs trained on $\mathcal{E}_V$ and $\mathcal{E}_{VT}$ in dog domain. Since the accuracy of $\mathcal{E}_V$ heavily relies on the confidence threshold $\alpha$ as showed in Figure~\ref{SLC}, we try three different settings with $\alpha\in\{0,5, 0.7, 0.9\}$ for constructing $\mathcal{E}_V^{dog}$ in this experiment. The experimental results in Table~\ref{without_semantic} show the performance of DCNN trained on $\mathcal{E}_V^{dog}\cup \mathcal{D}_{ImageNet}^{dog}$ is improved by increasing the confidence threshold since higher confidence threshold can lead to highly accurate dataset. However, the performance of DCNNs trained with $\mathcal{E}_V^{dog}$ is lower than the DCNN trained on $\mathcal{D}_{ImageNet}^{dog}$, which means that DCNN still cannot improve itself by self-labeling images without using contextual information from Web. 

\begin{table}
	\begin{center}
		\begin{tabular}{|l|c|c|c|}
			\hline
			\multirow{2}{*}{ $\mathcal{D}_{ImageNet}^{dog}$}& \multicolumn{3}{c|}{$\mathcal{E}_V^{dog}\cup \mathcal{D}_{ImageNet}^{dog}$}  \\\cline{2-4}
			& $\alpha = 0.5$ & $\alpha = 0.7$ &  $\alpha = 0.9$ \\\hline
			\hline
			65.80 & 64.22 & 65.30 & 65.53 \\
			\hline
		\end{tabular}
	\end{center}
	\caption{Single-crop top-1 accuracy of DCNNs trained on augmented datasets without using contextual information from Web.}\label{without_semantic}
\end{table}

We further investigate whether better model design and automatically labeled larger dataset can boost recognition performance together. Here, we choose ResNet-50~\cite{he2015deep} which performs much better than AlexNet on ILSVRC-2012 to repeat the experiment on the dog domain. Table~\ref{resnet_dog} reports the results, where ResNet-50 consistently outperforms AlexNet as expected, and ResNet-50 also improves itself using the automatically labeled data, which demonstrates that better model design and larger automatically labeled dataset can further boost the performance together.  % the visual restriction from a better performance DCNN is benefit to construct a higher quality dataset.

%To analysis how visual restriction from DCNN influence the quality of constructed dataset. We train a much complexer recognition model ResNet-50~\cite{he2015deep} instead of AlexNet on $\mathcal{D}_{ImageNet}^{dog}$ to label the candidate dataset $\mathcal{E}_T^{dog}$. Then we trained ResNet-50 on the human labeled dataset and the constructed dataset labeled by Web and ResNet-50. Table~\ref{resnet_dog} shows the test accuracy of these models. The experimental results show that the visual restriction from a better performance DCNN is benefit to construct a higher quality dataset.

\begin{table}\addtolength{\tabcolsep}{-1.5pt}
	\begin{center}
		\begin{tabular}{|l|c|c|c|}
			\hline
			\multirow{2}{*}{DCNN}& \multicolumn{3}{c|}{Training Data}  \\\cline{2-4}
			& $\mathcal{D}_{ImageNet}^{dog}$ &~~~$\mathcal{E}_{VT}^{dog}$~~~ & $\mathcal{E}_{VT}^{dog}\cup \mathcal{D}_{ImageNet}^{dog}$ \\\hline
			\hline
			AlexNet & 65.80 & 67.56 & \textbf{70.22} \\
			\hline
			ResNet-50 & 76.66 & 78.61 & \textbf{81.02} \\
			\hline
		\end{tabular}
	\end{center}
	\caption{Single-crop top-1 accuracy of different DCNNs trained on human labeled dataset and automatically labeled dataset.}\label{resnet_dog}
\end{table}

%We also curious about whether the dataset constructed labeled by low performance DCNN such as AlexNet could improve the performance of higher performance DCNN such as ResNet further or not. Thus we used the augmented dog dataset based on AlexNet to train a deeper model ResNet-50~\cite{he2015deep}. Table~\ref{resnet_dog} shows the recognition accuracy of ResNet-50 on Dog and Dog-A. The experimental results show that even only using the images constructed from Web by the low accuracy simple classifier AlexNet, it still have a significant improvement on the performance of much complexer recognition model. It demonstrated the inference described at section~\ref{jointly_restriction} that the quality of dataset constructed according to semantic-visual jointly restriction less dependent on the performance of recognition model learned from labeled data, and the dataset constructed by our proposed method contains more visual and semantic information than limited image labeled dataset, which can significant prevent overfitting and improve performance.

\subsection{Results on ILSVRC-2012}
We also try to augment ILSVRC-2012 training set ($\mathcal{D}_{ImageNet}^{1K}$) based on our proposed method. For categories with more than 15,000 images, we just keep 15,000 images by random sampling. After that there are 12.5 millions of images left in the augmented ILSVRC-2012, and most of the categories have more than 10,000 images, but there are still several rare categories contain fewer than 6000 images. Considering that unbalanced dataset for training can lead to poor performance since the validation set is a balanced one, then we balance the distribution of the augmented dataset by subsampling categories with more than 6,000 images, and construct a balanced dataset $\mathcal{E}_{VT}^{1K}$ with 5.7 million of images. 

The experimental results in Table~\ref{ilsvrc_res} show that the top-1 and top-5 classification accuracy on the validation set of ILSVRC-2012 with a single crop being evaluated. We found that classification performance to a large extent is affected by the number of training iterations. Models training on larger training dataset needs more iterations to be fully converged. Best performance is archived on both AlexNet and ResNet-50 by merging the human-label dataset and augmented dataset. It is worth noting that the augmented dataset $\mathcal{E}_{VT}^{1K}$ is labeled by a low performance AlexNet whose top-1 accuracy is 56.15\%, but the augmented dataset can still boost a high performance ResNet-50 from 74.55\% to 77.36\%. We also evaluated the performance of DCNN without dropout layers. The experimental results in Table~\ref{ilsvrc_res} shows that the DCNN without dropout layers can converge faster, and the influence of overfitting is alleviated and better performance is achieved thanks to the large scale augmented dataset. 

However, the performance by only using the automatically constructed is still lower than the human labeled dataset. We will do careful analysis in next section.
 %  single-crop top-1 accuracy of AlexNet trained on ILSVRC-2012 can be improved from 58.16\% to 59.41\% by using augmented dataset. %The performance should be improved further if using much more augmented images. We will report that experimental results on both of AlexNet and ResNet after we finished constructing the dataset.

\begin{table*}
	\begin{center}
		\begin{tabular}{|l|c|c|c|c|c|}\hline
			%\multirow{2}{*}{\#Iters}& \multicolumn{5}{c|}{Training Data  (Top-1 Accuracy)}  \\\cline{2-6}
			\multirow{2}{*}{DCNN}&\multirow{2}{*}{\#Iters}&\multirow{2}{*}{$\mathcal{D}_{ImageNet}^{1K}$} &\multirow{2}{*}{$\mathcal{E}_{VT}^{1K}$} & \multicolumn{2}{c|}{$\mathcal{E}_{VT}^{1K}\cup \mathcal{D}_{ImageNet}^{1K}$} \\\cline{5-6}
			& & & & Merge & Merge (No Dropout) \\\hline
			\hline
			\multirow{2}{*}{AlexNet}&0.4M & 56.15 (78.11) & 51.99 (73.86) & 56.13 (79.27) & 59.90 (81.17) \\
			\cline{2-6}
			&2.0M &  60.36 (82.38) & 56.58 (78.57) & \textbf{62.71 (83.71)} & 61.72 (82.62) \\
			\hline
			\hline
			\multirow{2}{*}{ResNet-50}& 0.5M & 74.55 (92.06) &  67.25 (85.99) & 75.57 (91.83) & -  \\
			\cline{2-6}
			&2.5M & 74.44 (92.11) & 70.17 (88.09) & \textbf{77.36 (93.29)}  & - \\
			\hline
		\end{tabular}
	\end{center}
	\caption{Single-crop top-1 (top-5) accuracy of AlexNet trained on human labeled datasets and augmented datasets.}\label{ilsvrc_res}
\end{table*}

\subsubsection{Dataset Analysis}\label{dataset_analysis}
\begin{figure}[!ht]
	\centering
	\begin{minipage}[t]{0.49\linewidth}
		\centering
		\includegraphics[page=27, width=1.0\linewidth]{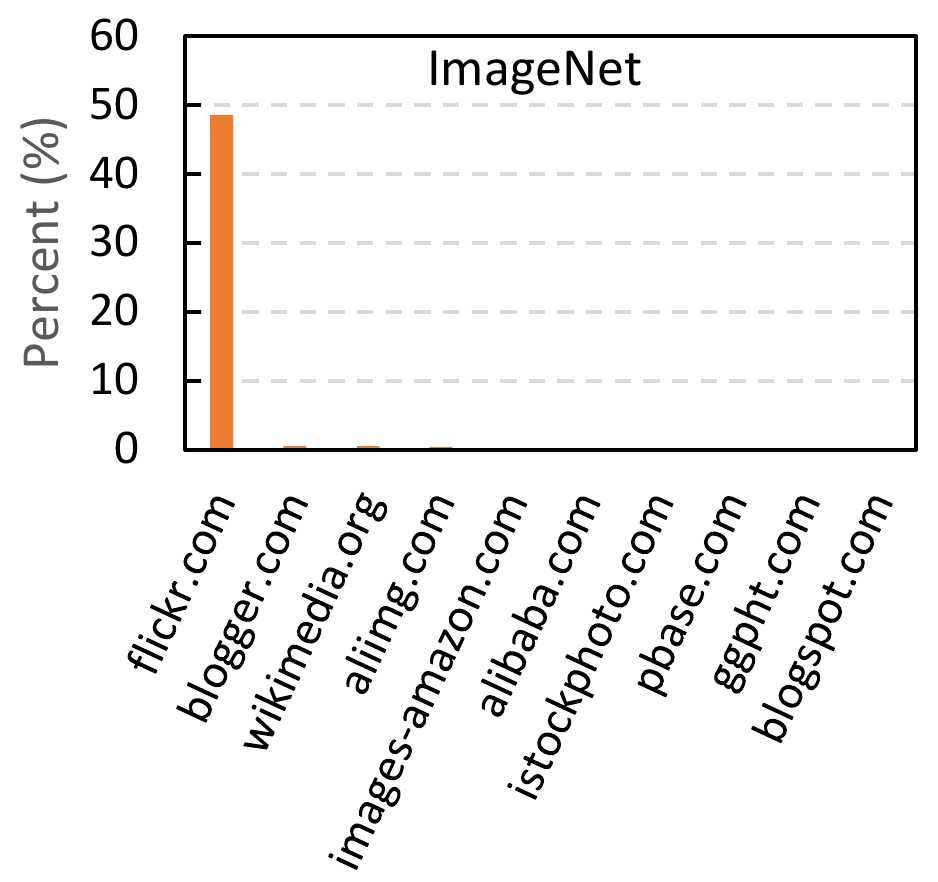}
	\end{minipage}~~%
	\begin{minipage}[t]{0.49\linewidth}
		\centering
		\includegraphics[page=28, width=1.0\linewidth]{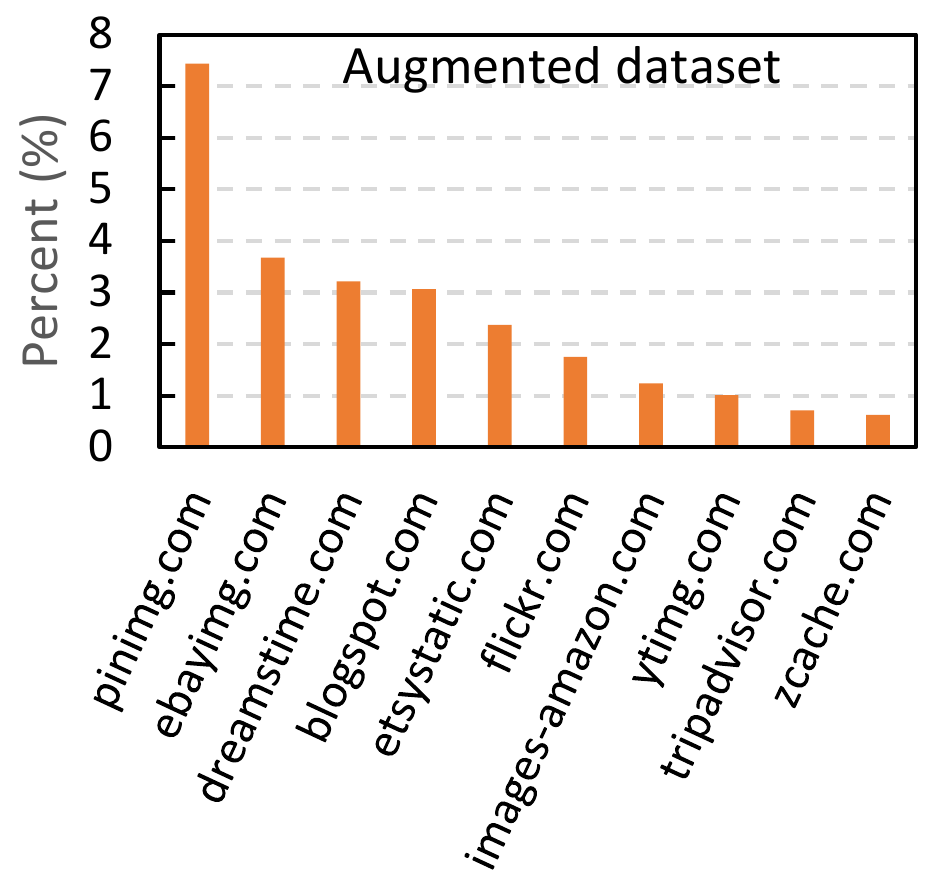}
	\end{minipage}%
	
	\caption{The distributions of top-10 frequent domains in human labeled datasets $\mathcal{D}_{ImageNet}$ and the automatically labeled datasets $\mathcal{E}_{VT}^{1K}$ respectively.}
	\label{domain_distributions}
\end{figure}

The performance gap comes from the distribution difference between the two datasets. ImageNet is collected about ten years ago where visual appearance of many categories are changed over time especially some man-made categories such as monitors and table lamp. In addition, the main source of ImageNet is Flickr, while our augmented dataset is from a wider range of websites where some are even not existing during ImageNet collecting such as pinterest. Figure~\ref{domain_distributions} shows the difference of domain distributions of image source of ImageNet and our augmented dataset respectively.

%We can notice that our augmented dataset $\mathcal{E}_{VT}^{1K}$ has lower performance than human labeled dataset, which is different with the experimental results reported in section~\ref{recognition_results}. There could be a possible bias between human labeled images and web collected images on the categories in ILSVRC-2012. As the work of Nizar et al.~\cite{massouh2017learning}
%indicated that, \textit{the perceptual object knowledge in ImageNet is static, its main body was collected during a limited and specific period (circa 2010~\footnote{http://image-net.org/about-stats}). This can result in some classes becoming dated over time, especially those representing man-made objects.} 

%Meanwhile, the ImageNet dataset collected from search engines, while our augmented dataset is collected from the whole web. Figure~\ref{domain_distributions} show the domain distributions of image source of ImageNet and our augmented dataset respectively. We found that nearly half of the images in ImageNet were collected from Flickr, whose image style is realistic rather than artistic, commercial, abstract, animated and so on. While the images in our augmented dataset are collected from all over the web and has better visual diversity. 

\begin{figure}[!ht]
	\centering
	\includegraphics[page=29, width=1.0\linewidth]{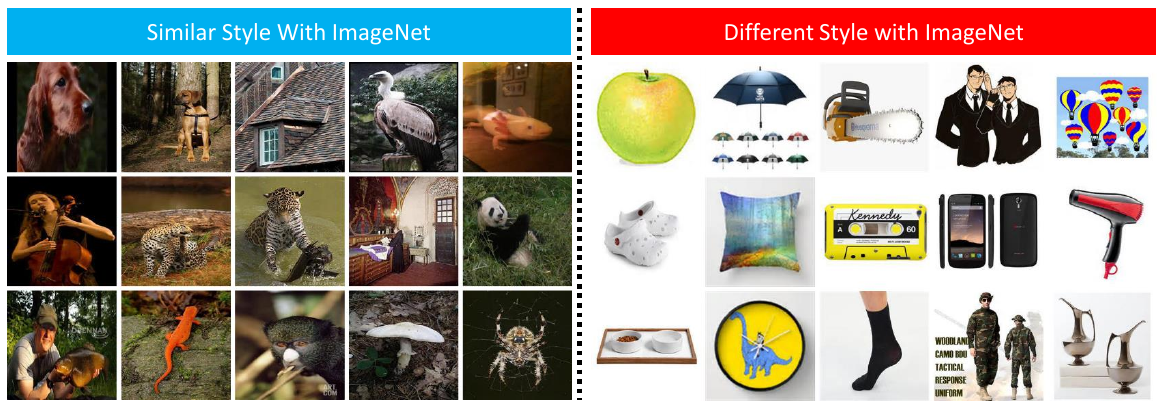}
	\caption{Images in $\mathcal{E}_{VT}^{1K}$ which are sorted by the output of $f_{critic}$. Images in the left figure are all with high value that means their image styles is closed to ImageNet, while images with low $f_{critic}$ outputs are shown in the right figure.}
	\label{different_style}
\end{figure}

To systematically study the distribution difference between the two datasets, we train the discriminator of Wasserstein Generative Adversarial Network (GAN)~\cite{Arjovsky2017Wasserstein} to differentiate images in ILSVRC-12 and images in our dataset by maximizing the Wasserstein distance between $\mathcal{D}_{ImageNet}$ and $\mathcal{E}_{VT}^{1K}$: $J_{critic} = \left [ \sum_{I_i \in \mathcal{D}_{ImageNet}}f_{critic}(I_i) - \sum_{I_i^{'} \in \mathcal{E}_{VT}^{1K}}f_{critic}(I_i^{'})\right ] $. By using the trained discriminator model $f_{critic}$, we sorted the images in  $\mathcal{E}_{VT}^{1K}$ according to the output value of $f_{critic}$ and show the images whose styles are most different/similar with  $\mathcal{D}_{ImageNet}$ in Figure~\ref{different_style}. We found that many images are easily distinguished from images in ILSVRC-2012 are collected from e-commence websites.

%Inspired from the discriminator of Wasserstein Generative Adversarial Networks, we tried to train a deep neural network $f_{critic}$ based on Wasserstein distance to distinguish the human labeled ILSVRC-2012 and our augmented dataset. The network contains five convolutional layers. the second, third and fourth convolutional layers are followed by batch normalization layer. The ReLu nonlinearity is applied to each hidden convolutional layer. Our optimization target is to maximize the Wasserstein distance between $\mathcal{D}_{ImageNet}$ and $\mathcal{E}_{VT}^{1K}$: $J_{critic} = \frac{1}{m}\left [ \sum_{i=1}^{m}f_{critic}(I_i) - \sum_{i=1}^{m}f_{critic}(I_i^{'})\right ] $, where $I_i \in \mathcal{D}_{ImageNet}$, $I_i^{'} \in \mathcal{E}_{VT}^{1K}$, and $m$ is size of min-batch. By using the trained discriminator model $f_{critic}$, we sorted the images in  $\mathcal{E}_{VT}^{1K}$ according to the output value of $f_{critic}$ and show the images whose styles are most different with the styles of images in $\mathcal{D}_{ImageNet}$ in Figure~\ref{different_style}. We found that the most of images which can be easily distinguished from images in ILSVRC-2012 are collected from shopping websites.%, then split the dataset into two different style： ImageNet style and Web style. 

Considering the difference between ImageNet and our dataset are mainly on man-made categories, we split the 1000 categories into two subsets according WordNet ontology, one is artifact set including 522 categories, the other is natural object set including 478 categories. We compare DCNNs trained on these two subset with ImageNet respectively, where Table~\ref{nature_artifact_recognition_performance} reports the performance. As expect, our dataset achieve better results on natural categories while worse on artificial categories. Since many images in ImageNet are out-of-date, we will do more evaluation to verify our dataset.

%%%% nature, artifact domain
%\begin{table}
%	\small
%	\begin{center}
%		\begin{tabular}{|l|c||l|c|}\hline
%			$\mathcal{D}_{ImageNet}$& Percent(\%) & $\mathcal{E}_{VT}$ & Percent(\%)  \\\hline
%			\hline
%			flickr.com & 48.64 & pinimg.com & 7.4384 \\
%			\hline
%			blogger.com & 0.4904 & ebayimg.com & 3.6732 \\
%			\hline
%			wikimedia.org& 	0.476	& 	dreamstime.com& 	3.2224\\
%			\hline
%			aliimg.com& 	0.4251	& 	blogspot.com& 	3.0659\\
%			\hline
%			images-amazon.com& 	0.257& 		etsystatic.com& 	2.3751\\
%			\hline
%			alibaba.com& 	0.168	& 	flickr.com& 	1.7504 \\
%			\hline
%			istockphoto.com& 	0.1598& 		images-amazon.com& 	1.2447\\
%			\hline
%			pbase.com& 	0.1523	& 	ytimg.com& 	1.015\\
%			\hline
%			ggpht.com& 	0.1086	& 	tripadvisor.com& 	0.7167\\
%			\hline
%			blogspot.com& 	0.1069	& 	zcache.com& 	0.633\\
%			\hline
%		\end{tabular}
%	\end{center}
%	\caption{Single-crop top-1 accuracy of DCNNs trained on human labeled datasets and augmented datasets.}\label{domain_distributions}
%\end{table}

%%% nature, artifact domain
\begin{table}
	\begin{center}
		\begin{tabular}{|l|c|c|c|c|}\hline
			\multirow{2}{*}{Domain}& \multirow{2}{*}{\#Iters} & \multicolumn{2}{c|}{Training Data}  \\\cline{3-4}
			&& $\mathcal{D}_{ImageNet}$ &~~~$\mathcal{E}_{VT}$~~~\\\hline
			\hline
			Natural & 2.0M & 68.17 & 69.53  \\
			\hline
			Artifact & 2.0M & 57.05 & 52.23 \\
			\hline
		\end{tabular}
	\end{center}
	\caption{Single-crop top-1 accuracy of DCNNs trained on human labeled datasets and augmented datasets.}\label{nature_artifact_recognition_performance}
\end{table}
  
%% domain analyssi

%% wgan analysis  
  
%We present some qualitative analysis on the bias between these two datasets in two different ways:
%
%\textbf{Data Sources.}
%
%\textbf{Image }

\subsubsection{Results on Cross-Dataset Generalization}
To further compare ILSVRC-2012 and our constructed dataset, we verify the cross-dataset generalization ability of these two datasets. Cross-dataset generalization measures
the performance of classifiers learned from one dataset on the other dataset. We compare our augmented dataset with WebVision~\cite{webvison} dataset which is constructed from Flickr and Google Images Search by querying the category names. Table~\ref{cross_dataset} shows the classification error rates. Each dataset produces a DCNN using its training set, and then evaluate the trained model on test set from different dataset. In all of the cases, the best performance is achieved
by training and testing on the same dataset. The experimental results shows that our augmented dataset have better performance than ImageNet on WebVision. Moreover, our augmented dataset also achieves better performance than WebVision on human labeled image dataset ImageNet. Overall, our dataset generalizes much better than the other two datasets.

%\begin{table}
%	\begin{center}
%		\begin{tabular}{|l|c|c|}\hline
%			\multirow{2}{*}{Training Data}& \multicolumn{2}{c|}{Test Data}  \\\cline{2-3}
%			& ILSVRC 2012 Val & WebVision Val \\\hline
%			\hline
%			$\mathcal{D}_{ImageNet}$ & 56.15 (78.11) & 52.58 (74.64)  \\
%			\hline
%			WebVision & 47.55 (70.36) & 57.03 (77.90) \\
%			\hline
%			$\mathcal{E}_{VT}^{1K}$ & 51.99 (73.86) & 53.54 (73.86)  \\
%			\hline
%			\hline
%			$\mathcal{D}_{ImageNet}$+ & 60.36 (82.38) & 54.99 (76.63) \\
%			\hline
%			$\mathcal{E}_{VT}^{1K}$+ & 56.58 (78.57) & 57.98 (77.45)\\
%			\hline
%		\end{tabular}
%	\end{center}
%	\caption{Top-1 accuracy of DCNNs trained on human labeled datasets and augmented datasets by using dense test.}\label{cross_dataset}
%\end{table}

\begin{table}
	\begin{center}
		\begin{tabular}{|l|c|c|}\hline
			\multirow{2}{*}{Training Data}& \multicolumn{2}{c|}{Test Data}  \\\cline{2-3}
			& ILSVRC 2012 Val & WebVision Val \\\hline
			\hline
			$\mathcal{D}_{ImageNet}^{1K}$ & 56.15 & 52.58  \\
			\hline
			WebVision & 47.55 & 57.03 \\
			\hline
			$\mathcal{E}_{VT}^{1K}$ & 51.99 & 53.94 \\
			\hline
			\hline
			$\mathcal{D}_{ImageNet}^{1K}$* & 60.36 & 54.99 \\
			\hline
			$\mathcal{E}_{VT}^{1K}$* & 56.58 & 57.98 \\
			\hline
		\end{tabular}
	\end{center}
	\caption{Top-1 accuracy of DCNNs trained on human labeled datasets and augmented datasets by using dense test. The experimental results with mark $*$ are trained with 2.0M iterations, the others are trained with 0.4M iterations.}\label{cross_dataset}
\end{table}

\subsubsection{Results of MSR-Bing Grand Challenge}
Inspired by the success of feature extractors in DCNNs learned from ILSVRC-2012, we also try to compare the generalization ability of features extractors learned from human labeled ILSVRC-2012 and our augmented ILSVRC-2012 dataset. To evaluate the quality of feature extractors in a more comprehensive way, we test the performance of the feature extractors on an open domain image retrieval task - MSR-Bing Grand Challenge~\cite{hua2013clickage}. %and PASCAL VOC object classification task.

The MSR-Bing Grand Challenge task provides a training set including 11.7 million queries and 1 million images, a test set including 1000 queries and 79,665 images, and requires to learn a ranking model based on training set and rank images for each query in test set, where Normalized Discounted Cumulative Gain ($NDCG$) is used as evaluation metric for a ranking list, which is defined as
\begin{equation}
NDCG@d = Z_d\sum_{j=1}^d\frac{2^{r^j}-1}{log(1+j)}
\end{equation}
where $r^j$ = {Excellent = 3, Good = 2, Bad = 0} is the manually judged relevance for an image ranked at $j$ with respect to the query, $Z_d$ is a normalization factor to make the score to be 1 for ideal case. The performance is measured by average $NDCG@d$ on all queries in test set.

\begin{figure}[ht]
	\centering
	\includegraphics[width=0.47\textwidth,page=18]{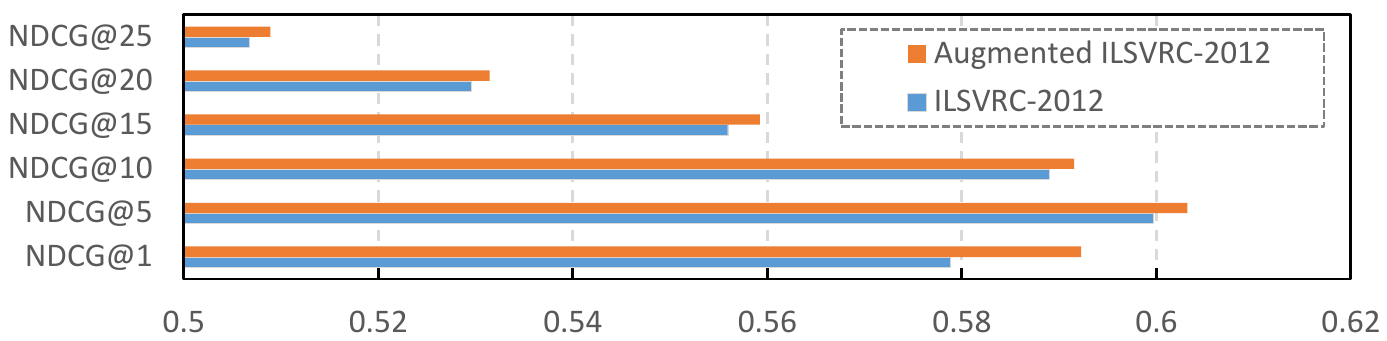}
	\caption{The $NDCG$ of CCA for image
		search using image representations provided by DCNNs trained on ILSVRC-2012 training set and augmented ILSVCR-2012 training set.}\label{ndcg_distributions}
\end{figure}

We use Canonical Correlation Analysis (CCA)~\cite{cca} as the basic ranking model and represent a query with bag-of-textual-words. For images, we use the outputs of the last but one fully-connected layer of a DCNN as the image representation, and two DCNNs trained on ILSVRC-2012 and augmented ILSVRC-2012 will be used. Figure~\ref{ndcg_distributions} compares the performance of ranking results using image representations provided by the two DCNNs, where the DCNN trained on augmented ILSVRC-2012 achieves consistently better performance, which further demonstrates the generalization ability of model learned from automatically labeled dataset.

\section{Conclusion}\label{conclusion}
In this paper, we propose a method to do automatic dataset augmentation, where both Web and DCNN are used. 
Specifically, Web provides massive images with rich contextual information, while well-trained DCNNs are used to label these images and filter out noisy images. Meanwhile, the rich contextual information from Web ensures DCNN to achieve high labeling accuracy with relatively low confidence threshold. Together, we can augment an labeled image dataset in a scalable, accurate, and informative way.
Extensive experiments demonstrate that well-trained DCNNs can automatically label images from Web and further improves themselves with the automatically labeled datasets. We hope the automatically constructed large-scale datasets with rich contextual information can help research in large neural networks.

\bibliographystyle{abbrv}
\bibliography{reference}

\end{document}